\documentclass[conference,compsoc]{IEEEtran}

\usepackage{amsmath,amssymb,amsfonts}
\usepackage{adjustbox}
\usepackage{algorithmic, algorithm}
\usepackage{hyperref}  
\usepackage{multirow}
\usepackage{graphicx}
\usepackage{subfigure}

\usepackage{textcomp}
\usepackage{xcolor}
\def\BibTeX{{\rm B\kern-.05em{\sc i\kern-.025em b}\kern-.08em
    T\kern-.1667em\lower.7ex\hbox{E}\kern-.125emX}}

\begin{document}%

\title{Extract Dynamic Information To Improve Time Series Modeling: \\a
  Case Study with Scientific Workflow}

\author{\IEEEauthorblockN{Jeeyung Kim\IEEEauthorrefmark{1},
                          Mengtian Jin\IEEEauthorrefmark{2},
                          Youkow Homma\IEEEauthorrefmark{2},
                          Alex Sim\IEEEauthorrefmark{1},
                          Wilko Kroeger\IEEEauthorrefmark{3},
                          Kesheng Wu\IEEEauthorrefmark{1}}\\
\IEEEauthorblockA{\IEEEauthorrefmark{1}\textit{Lawrence Berkeley National Laboratory, Berkeley, CA, USA} \\
\{jeeyungkim, asim, kwu\}@lbl.gov}
\IEEEauthorblockA{\IEEEauthorrefmark{2}
\textit{Stanford University, Stanford, CA, USA}\\
\{mtjin, yhomma\}@stanford.edu}
\IEEEauthorblockA{\IEEEauthorrefmark{3}
\textit{SLAC National Accelerator Laboratory, Stanford, CA, USA}\\
wilko@slac.stanford.edu}
}

\maketitle

\begin{abstract}
 In modeling time series data, we often need to augment the existing
 data records to increase the modeling accuracy.
 In this work, we describe a number of techniques to extract dynamic information
 about the current state of a large scientific workflow,
 which could be generalized to other types of applications.
 The specific task to be modeled is the time needed for
 transferring a file from an experimental facility to a data center.
 The key idea of our approach is to find recent past data transfer
 events that match the current event in some ways.
 Tests showed that we could identify recent events matching
 some recorded properties and reduce the prediction error by about 12\%
 compared to the similar models with only static features.
 We additionally explored an application specific technique to extract information
 about the data production process, 
 and was able to reduce the average prediction error by 44\%.
\end{abstract}

\section{Introduction}
\label{sec:intro}
Automated data collection and analysis has become the key tool to
understand the world~\cite{Hey:2009:DSD, Kitchin:2014:DataRevolution}.
We frequently encounter measurements in time, such as evolution of
supernova~\cite{Cao:2013:DPE} and gravitational
wave~\cite{George:2018:DLR}.
In these cases, we often need to engineer new features to augment the
existing data records in order to improve the modeling accuracy, for
example, by including information from the most recent past data
record~\cite{Domingos:2012:AML, Zheng:2018:FEM}.
In general, feature engineering is a critical step in any data analysis
tasks.
Specific tasks for feature engineering could include organizing data
into a convenient for the analysis platform, filling in the missing
values, removing outliers, normalizing the values, as well as the above
mentioned augmentation of data records.
In this work, we explore a number of techniques to extract dynamic
information to improve the prediction accuracy for a specific scientific
workflow (to be described in more details in
Section~\ref{sec:testcase}).
However, we believe the majority of the techniques are easily
applicable to other types of time series modeling efforts.


Time series data is a common output from automated data acquisition
systems~\cite{Bleikh:2016:TSA, deGooijer:1992:TSM,
  Mahalakshmi:2016:SFT}.
We broadly regard any data collection where each record contains a time
stamp as a time series.
This time stamp feature allows us to define a notion of time and compare
data records according to their time stamps.
Given any particular data record, we typically describe it as denoting
the \emph{current} event, a data record with a smaller value for its
time stamp as an event in the past, and a data record with a larger
time stamp value as an event in the future.
A common goal of modeling a time series is to use the information about
past events make predictions about the current event or
future events.
In this work we study the performance of transferring data
from an experimental facility to a data center.
We assume the performance prediction is made before the start of a file
transfer, where the available information include all information about
completed file transfers plus the static information already known about
the current file to be transferred.
How to best utilize the information about past information to improve
the prediction accuracy is the main concern of this work.

Data scientists often spends a large fraction of their working day
performing feature engineering tasks.
Due to a large variety of possible tasks and lack of guidance on how to
make effective choices, data scientists typically has to slowly and
manually sift through the myriad of choices.
There are a number of attempts to automate part of this process.
For example, \texttt{Brainwash} is designed to automate many of the routine work and
help data scientists to make choices more effectively~\cite{Anderson:2013:Brainwash}.
Some work focuses on addressing feature engineering challenges in
specific allocation domains~\cite{Chen:2019:IPM, Fan:2019:DLF}.
While others advocate for using deep learning to automate the selection
of features~\cite{Bang:2020:HWC, Fan:2019:DLF, Parvandeh:2020:CFN}.
In this work, we take advantage of the time stamps in data to develop
techniques for time series data.  These techniques can extract
information from past history to approximate important relevant states
of the system context not explicit captured by the data set.
Tests show that our approach is quite effective in reducing the
prediction errors.

Modeling of time series data also has its own unique challenges.
To determine the best model to fit a data set, one
typically randomly select a sample of the data for training, testing and
validation.
This random shuffling of data is critical to avoid overfitting in
selecting the appropriate modeling techniques and their associated
parameters~\cite{Arlot:2010:SCV, Bailey:2015:SOB}.
However, this random shuffle is not appropriate for time series data
because it does not preserve the time order.
To avoid building a model on future data to test the past observations,
we develop a nested cross validation~\cite{Bergmeir:2012:CVT,
  Cawley:2010:OMS, Wainer:2018:NCV} approach that preserves time order.

In summary, the key contributions of this work includes:
\begin{enumerate}
  \item Develop a set of techniques to extract dynamic system
    information from recent past data records.
  \item Design a time-order-preserving nested cross validation technique
    to determine the hyperparameters for the machine learning techniques
    examined.
  \item Test our algorithms with a large set of performance monitoring
    data from a large scientific workflow.  Tests show that our approach
    can significantly reduce the prediction errors.
\end{enumerate}


This paper is partly based on the work previously reported by Yang et al.~\cite{ylksw}.
Their work primarily focused on demonstrating the capability to predict
the data transfer performance for the application scientists.
This paper focuses on the lessons we could learn on feature engineering.
Furthermore, we have also conducted extensive tests with neural networks
to automate the feature extraction process.

In the remaining of this paper, we give a description of the scientific
workflow used as the test case in Section~\ref{sec:testcase}.  We
describe the efforts to augment the data records with recently completed
events in Sections~\ref{sec:lags-trees} and \ref{sec:lags-nn}, describe
the work of more precise dynamic information in Section~\ref{sec:runs}.
A brief summary of lessons learned is given in
Section~\ref{sec:summary}.

\section{LCLS:  a science use case}
\label{sec:testcase}
\begin{figure}
    \centering
    \includegraphics[width=\linewidth]{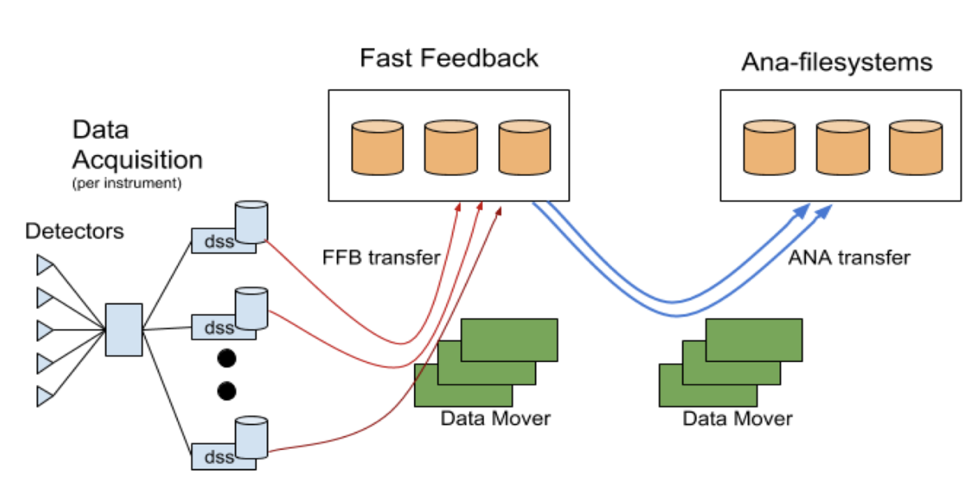}
    \caption{LCLS Data flow: Red line is DSS $\rightarrow$ FFB transfer, Blue line is FFB $\rightarrow$ ANA transfer}
    \label{fig:data_flow}
\end{figure}

Our example time series come from monitoring the data transfer operations between the Linac Coherent Light Source (LCLS) and the offline compute facility both located at the SLAC, National Accelerator Laboratory.
LCLS is a X-ray laser to create images of molecules, it is a basic tools
for chemistry, biology, physics and materials.
There are seven instruments with different detectors and X-ray beam characteristics that allow diverse types of experiments~\cite{ylksw}.
LCLS produces terabytes of data for each experiment
The current data pipeline involves two stages, the data files are
temporarily stored at Fast Feedback (FFB) system, close to the experiment, and then
to Analysis (ANA) system in the SLAC computing center.
FFB provides fast, low latency access to the collected data.
It is only used by the active experiments.
ANA is large in size (4PB), shared between all experiments and holds the
experimental data for many months.
The data flow in shown in Figure~\ref{fig:data_flow}.

When data is generated from an experiment, it is written to files on Data Storage Subnet (DSS)
nodes via the data acquisition (DAQ) system.
Multiple DSS nodes are used in parallel for each data collection (a run)
and files from the same node constitute a stream.
Typically 5-6 streams are run in parallel.
A data mover then transfers files from DSS to FFB.
The data is then transferred from FFB to ANA.
The typical FFB transfer rate is limited by the rate the file is written
to a DSS node, usually around 100-200MB/s.
The typical limit for the ANA transfer rate in this process is about 350-400MB/s due to the speed of checksum calculation.

In this work, our aim is to use historical data from the LCLS workflow
to build a model that can predict transfer rates of future transfers
from DSS nodes to FFB and FFB to ANA.
In particular, we conduct our analysis and model-building on a set of
258,765 transfers from May 2017 to January 2018 where 131,274 transfers
take place from DSS to FFB and the remaining 127,491 are from FFB to ANA.

\begin{table}
\centering
\begin{tabular}{|l|r|}
\hline
\textbf{Instrument}     & \textbf{Count} \\ \hline
cxi         & 28900      \\ \hline
xpp         & 23110      \\ \hline
mec         & 22509       \\ \hline
xcs         & 16307       \\ \hline
sxr         & 12354       \\ \hline
mfx         & 9713       \\ \hline
amo         & 8341      \\ \hline
\end{tabular}
\vspace{0.2cm}
\caption{Number of Records Collected by Each Instrument}
\label{table:instr_count}
\end{table}

\begin{table*}
\centering
\begin{tabular}{|l|l|l|l|}
\hline
Instrument & FFB File System & DSS $\longrightarrow$ FFB Host & FFB $\longrightarrow$ ANA Host \\ \hline
amo, sxr, xpp & ffb11 & psana\{102,103\} & psana\{102,103\}, psexport\{01,02,05,06,07,08\} \\ \hline
cxi, mec, mfx, xcs & ffb21 & psana\{201,202,203\} & psana\{201,203\}, psexport\{01,02,05,06,07,08\} \\ \hline
\end{tabular}
\vspace{0.2cm}
\caption{Summary of LCLS detectors: instruments in the top row are
  located in Near Experiment Hall (NEH) and those in the bottom row are
  located in the Far Experiment Hall (FEH).}
\label{table:instr}
\end{table*}

\begin{figure}
    \centering
    \includegraphics[scale=0.45]{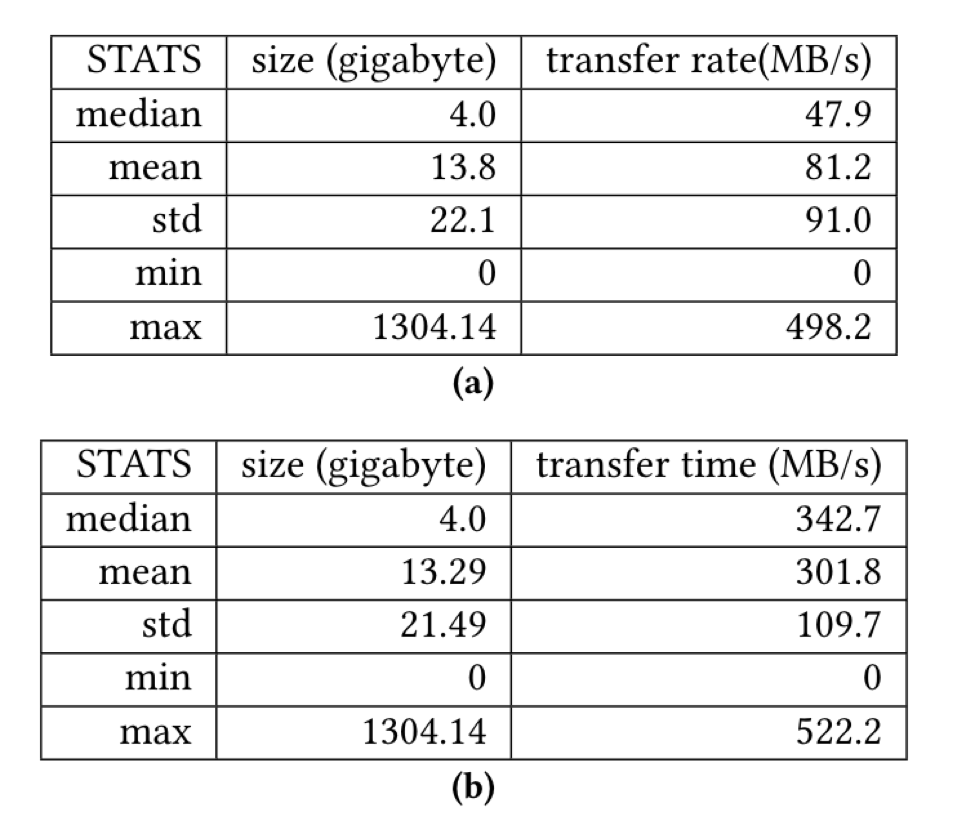}
    \caption{Basic Statistics Summary of Transfer Rate and File Size on FFB and ANA. (a) DSS to FFB (b) FFB to ANA: 12 records which have file sizes over 1 TB due to configuration errors, and 76 transfers with either a file size or transfer rate of zero are removed from the analysis.}
    \label{fig:frate_stats}
\end{figure}

LCLS has seven instrumental stations with different types of
detectors. The stations are distributed in two buildings known as the
Near Experiment Hall (NEH) and the Far Experiment Hall (FEH), see
Table~\ref{table:instr}.
Due to this physical separation, these instruments are also attached to
different local file systems.
Table~\ref{table:instr_count} has the number of files produced by each of
the seven instruments used in the data collection.
From this table we see that cxi, xpp and mec are used much more  than the others.
The overview of statistics on transfer rate and file size in FFB and ANA transfers are shown in Figure~\ref{fig:frate_stats}. 

Next, we describe the file generation and transport
process to give more information about what features are important to
predicting the file transfer performance.

\begin{figure}
    \centering
\begin{tabular}{cc}
  \includegraphics[width=1.5in]{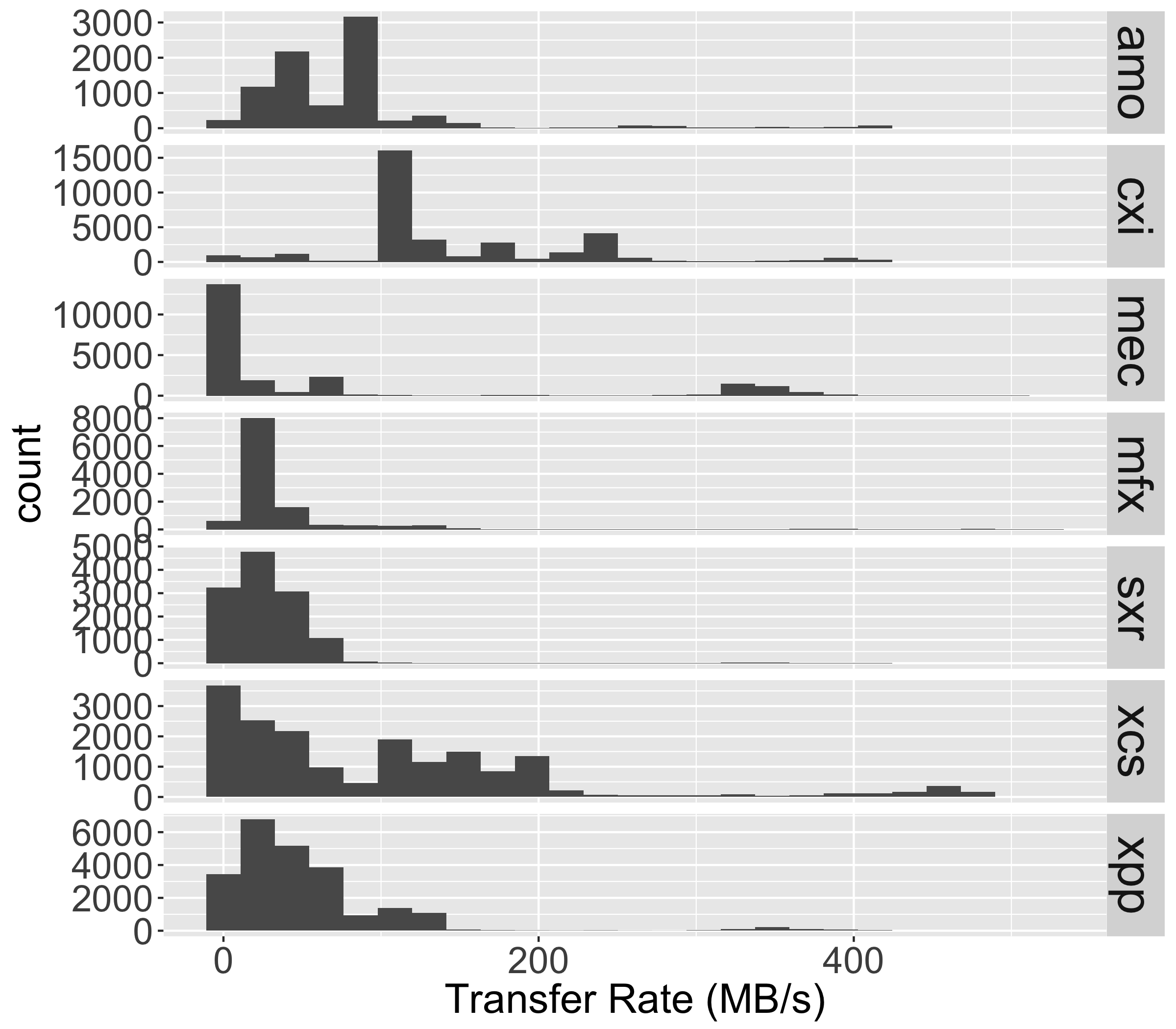}
  &
    \includegraphics[width=1.5in]{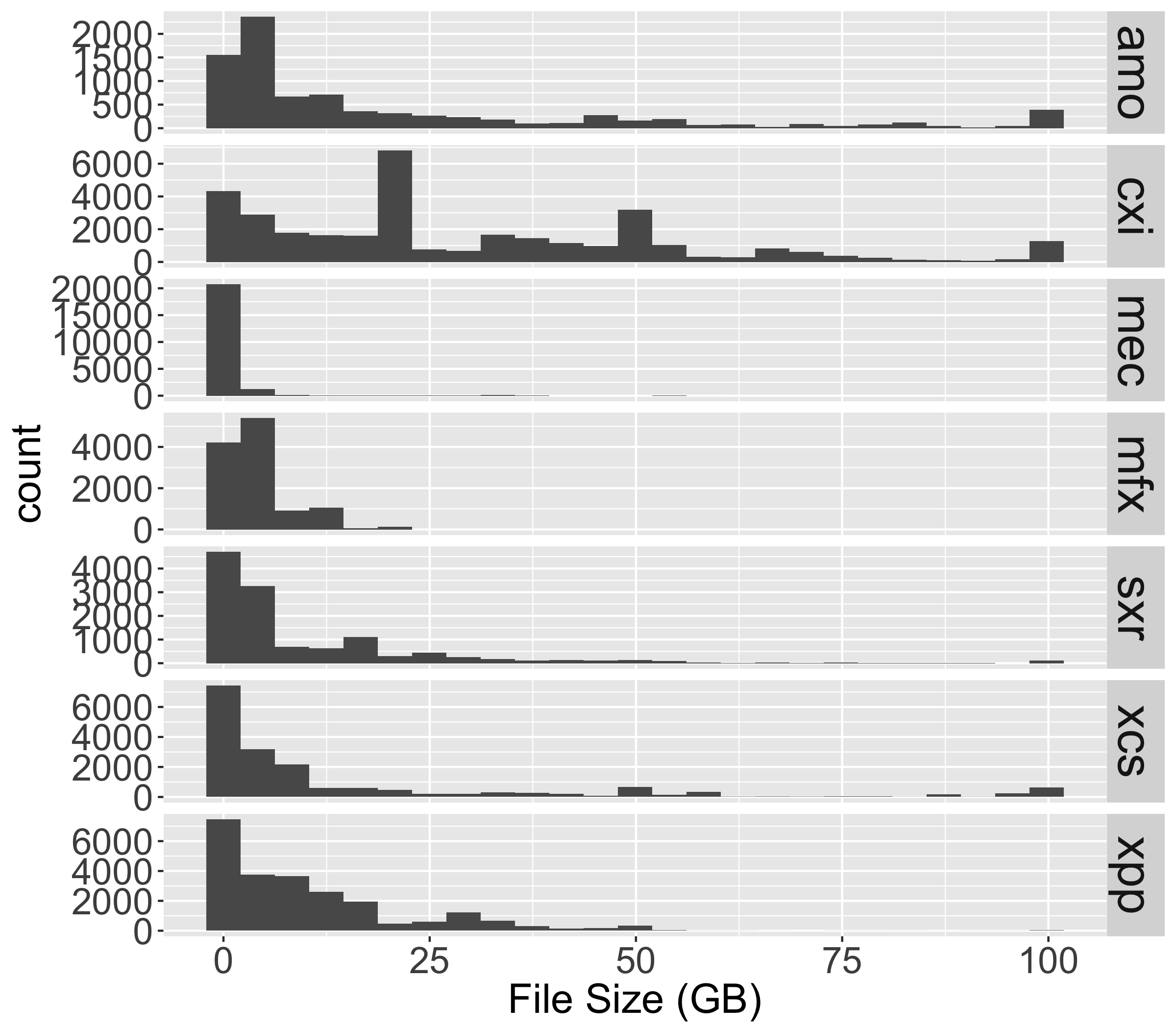}
  \\
  \parbox[t]{1.5in}{\scriptsize (a) Histogram of transfer rate (to FFB)}
  &
    \parbox[t]{1.5in}{\centering \scriptsize (b) Histogram of file size}
  \\
  \end{tabular}
    \caption{On the whole, 'cxi' and 'amo' tend to have faster
      transfer rates, while 'mec' has the slowest transfer rates.}
    \label{fig:ffb-instr-frate}
\end{figure}

\paragraph{File Size}
Based on the past experience, the performance of a file
transfer is heavily dependent on the file size.  The first feature we
plan to explore is the file size.
Figure \ref{fig:ffb-instr-frate} shows the distribution of the
file sizes and transfer rates.  We see that different instruments have
clearly distinctive distributions.  

\paragraph{Nodes and Streams}
As data is generated from an experiment, it is written in parallel
streams with each stream assigned to one DSS node.  This data is then
simultaneously written to the FFB by the data mover/host.  
Once a file for a stream reaches a cap of 100GB, which is configurable, the file is closed and a new file for the same stream is opened. We call a set of files for the same stream chunks.
The structure of chunks and streams are embedded in the file name and have to be
extracted.  Such information allows us to better understand the data
transport process and make more accurate predictions.

\begin{figure}
    \centering
    \includegraphics[width=3.25in]{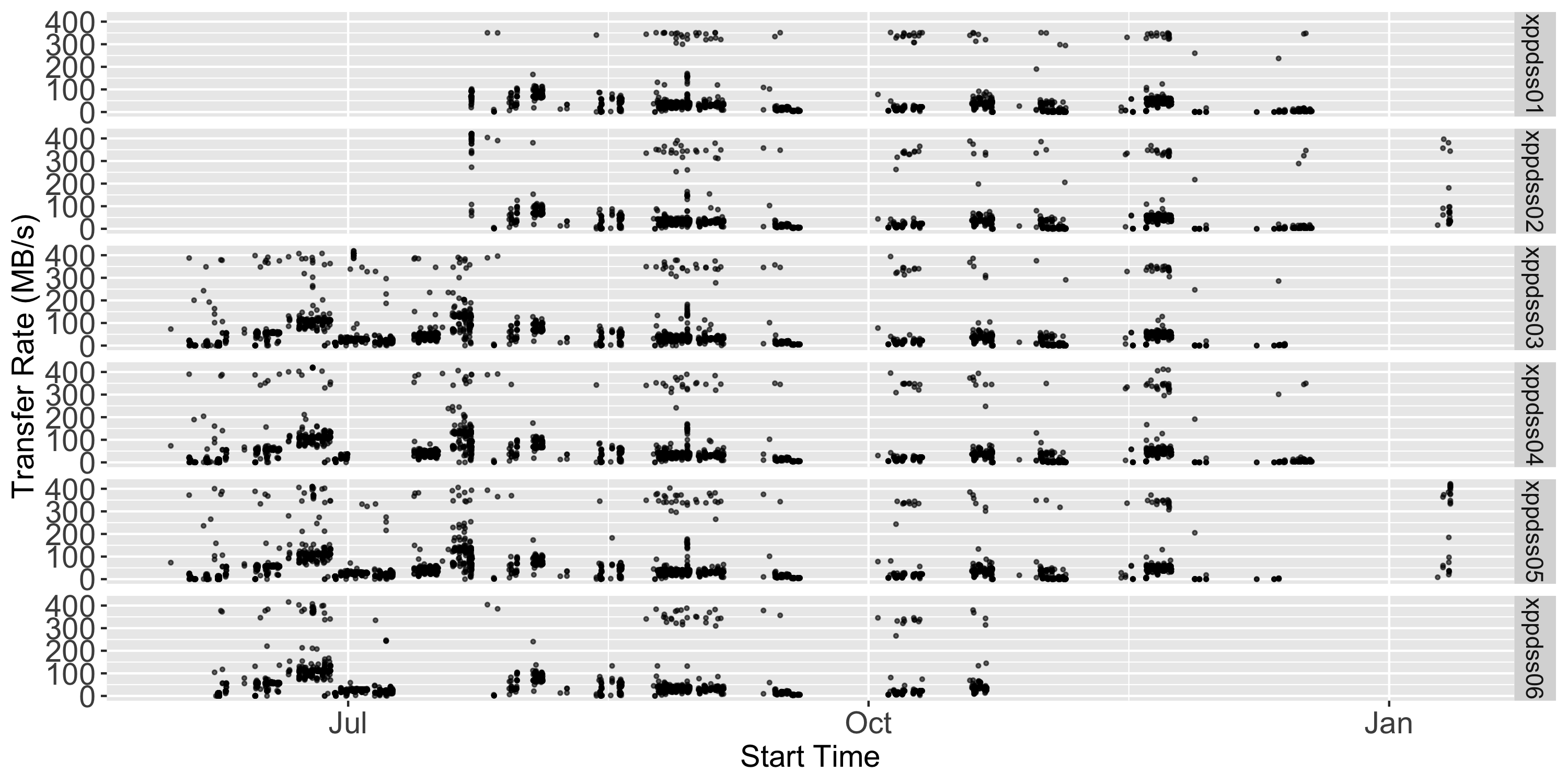}
    \caption{Scatter plot of start time vs transfer rate split by node for experiments on 'xpp'.}
    \label{fig:nodes-yearly}
\end{figure}

In Figure \ref{fig:nodes-yearly}, we plot the transfer rate over time for all nodes of instrument 'xpp,' and we can see that different nodes are dormant and active during different parts of the year.

\begin{figure}
    \centering
    \includegraphics[scale=0.3]{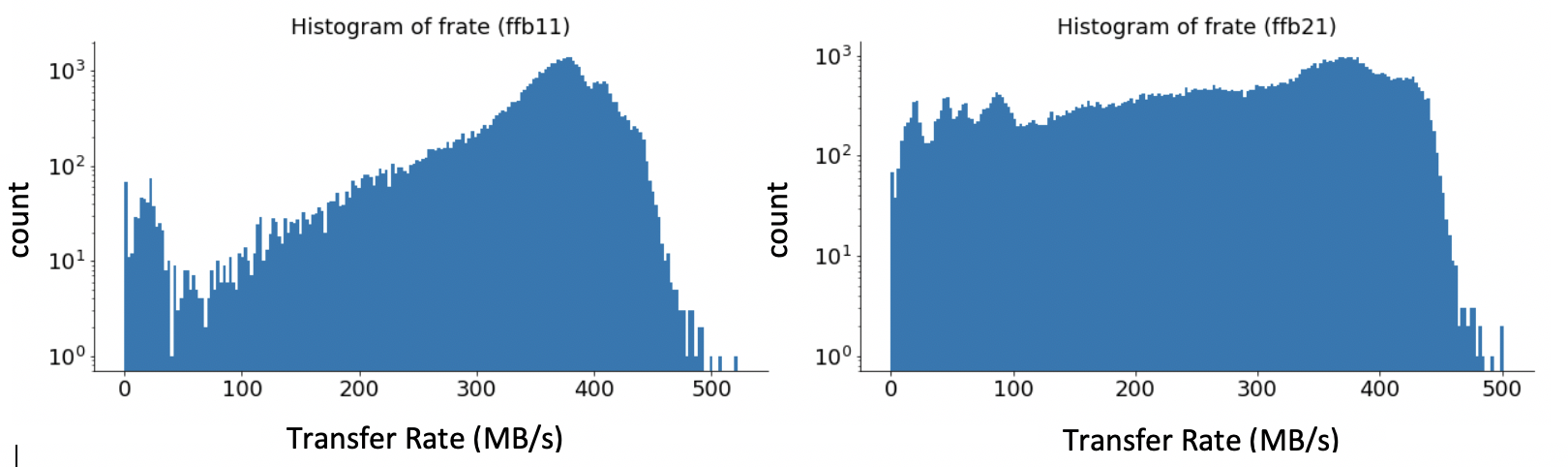}
    \caption{Distribution of transfer rates are different for different file systems.}
    \label{fig:sample_distribution}
\end{figure}

\paragraph{File systems}
From the data overview in Table~\ref{table:instr}, we also notice the
close relationship between instrument and FFB file system, which is the
source file system in the ANA transfer process. Therefore, we expect
this source file system also affects the transfer rate, and we show the
distribution plots of the two different source file systems in
Figure~\ref{fig:sample_distribution}.

\begin{table*}
\centering
\begin{tabular}{|l|l|r|r|l|l|}
\hline
Start Time          & Stop Time           & File Size (GB) & Transfer Rate (MB/s) & Target Host & Node     \\ \hline
2017-06-21 16:02:02 & 2017-06-21 16:02:26 & 0.3168954      & 13.73635             & psana201    & mfxdss02 \\ \hline
2017-06-21 16:02:02 & 2017-06-21 16:02:26 & 0.3168954      & 13.73635             & psana201    & mfxdss01 \\ \hline
2017-06-21 18:48:26 & 2017-06-21 18:48:28 & 0.3168954      & 399.08803            & psana203    & mfxdss05 \\ \hline
2017-06-21 18:48:26 & 2017-06-21 18:48:28 & 0.3168954      & 399.08803            & psana203    & mfxdss06 \\ \hline
\end{tabular}
\vspace{0.2cm}
\caption{Two streams from the same chunk are transferred more than  two
  hours later than other streams, which seems to cause significant
  difference in transfer rates.}
\label{table:delayed1}
\end{table*}

\begin{table*}
\centering
\begin{tabular}{|l|l|r|r|l|l|}
\hline
Start Time          & Stop Time           & File Size (GB) & Transfer Rate (MB/s) & Target Host & Node     \\ \hline
2017-09-22 19:48:26 & 2017-09-22 19:56:57 & 98.79810       & 193.4623             & psana201    & xcsdss03 \\ \hline
2017-09-22 19:48:26 & 2017-09-22 19:56:57 & 98.80620       & 193.7768             & psana201    & xcsdss02 \\ \hline
2017-09-22 19:50:42 & 2017-09-22 19:56:58 & 98.80619       & 263.1926             & psana202    & xcsdss05 \\ \hline
2017-09-22 19:53:31 & 2017-09-22 19:59:45 & 98.70099       & 264.1363             & psana202    & xcsdss06 \\ \hline
2017-09-22 19:55:07 & 2017-09-22 20:01:01 & 98.81428       & 279.6552             & psana202    & xcsdss04 \\ \hline
\end{tabular}
\vspace{0.2cm}
\caption{The last stream of this chunk has a minor delay of about two
  minutes, which seems to have a smaller impact on transfer rate.}
\label{table:delayed2}
\end{table*}

\paragraph{Delayed Transfers}
Different streams of a chunk are transferred at the same time, but there
are instances where a stream's start time is significantly later than
the start time of the first stream in the same chunk.
One example of this is in Table~\ref{table:delayed1} from chunk 00 of experiment 'e991-r0002' on instrument 'mfx'.  We see that the first two transfers are almost 3 hours before the later two streams.  The difference in target hosts suggests that there may have been a problem with psana203 during this time.
A slightly more nuanced example is in Table~\ref{table:delayed2}.  We see that the first two streams start at the same time, then there is about a 2 minute gap in subsequent streams.  This leads to a slight bump in transfer rate.  Similar to above, we see that the target host is different between the first two streams and the latter three.
 
\begin{figure}
    \centering
    \includegraphics[width=1.6in]{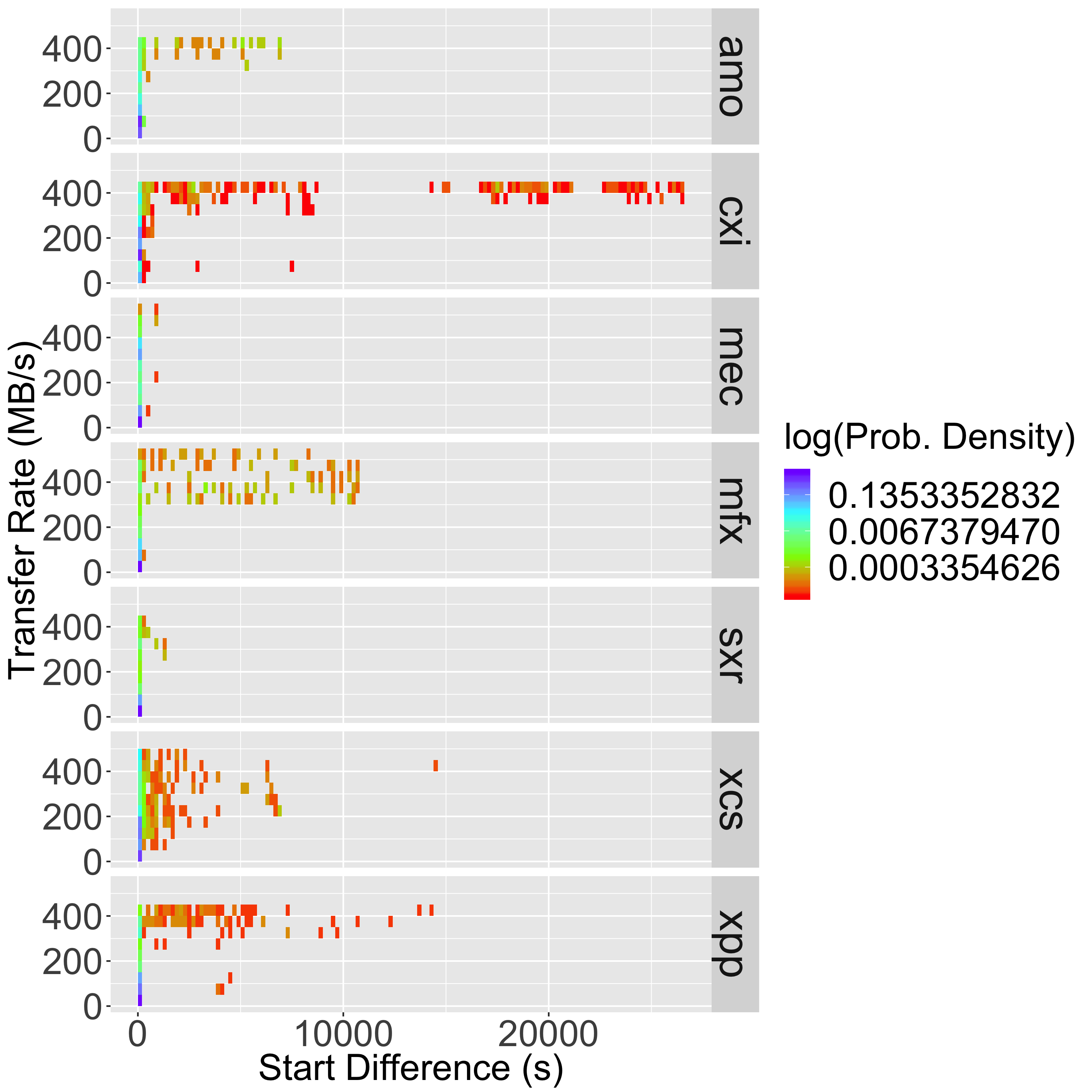}
    \includegraphics[width=1.6in]{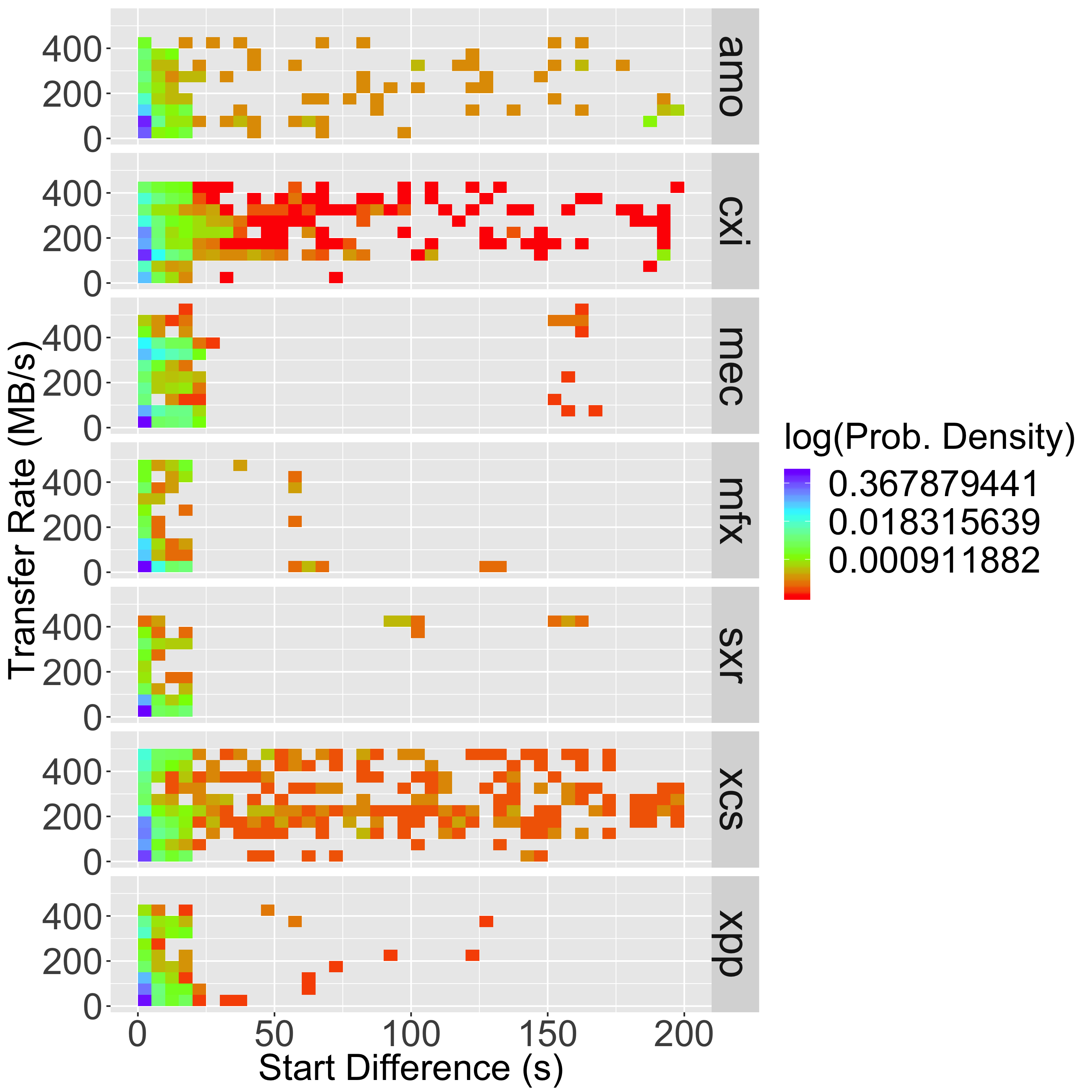}
    \caption{Impact of the start time difference on transfer rate. Left: Log density plot of start time difference vs transfer rate for all files. Right: Log density plot of start time difference vs transfer rate with time difference less than 200 seconds.}
    \label{fig:startt-rate}
\end{figure}

From Figure~\ref{fig:startt-rate}, we clearly see the impact of the
start time difference, especially when the start time difference is
large.
On the other hand, if the difference is on the order of a few minutes, then the pattern is less clear.

\section{Modeling LCLS Data Transfer Performance}
\label{sec:modeling}
In this section, we briefly review the main modeling approaches we will
be using to evaluate our feature engineering methods.  In addition, we
will also outline our cross validation procedure that respects time
order of the time series data, and establish a base line for evaluating
the effectiveness of the feature engineering efforts.

There is a significant amount of published work on modeling time series
data, we refer to the interested readers to books and reviews on this
subject matter for a comprehensive treatment~\cite{Bleikh:2016:TSA,
  Jensen:2017:TSM, Langkvist:2014:RUF, Mahalakshmi:2016:SFT,
  Sapankevych:2009:TSP}.
For this work, we have chosen to use a few common decision tree based
modeling approaches because they have high quality software
implementations that are freely available to us, and are efficient to
compute.  Additionally, they are known to be
effective in prediction, to produce interpretable results, and to
require a relatively small amount of hyperparameter
tuning~\cite{Galicia:2019:MSF, Kane:2014:CAR, Zhang:2015:GBM}.
In particular, we have selected to use an implementation of random
forest~\cite{rf} and a number of gradient boosting techniques~\cite{gbm, xgboost}.
We will also be exploring a few variantions of neural networks in
Section~\ref{sec:lags-nn}.
Despite their popularity, we found there are simply too many ways to
construct different neural networks and we have not been able to find
the best variant for our prediction task.
For this reason, we will be mostly staying with decision tree based
prediction models.

\paragraph{Order-preserving Nested Cross Validation}
For each modeling method, we tune the hyperparameters using nested cross validation (CV)
for time series, as described in Algorithm ~\ref{alg:cv}.  This nested
CV preserves the time order of the data records and prevents information leaking from the training set to the
validation set.
It guarantees this by having each element in the training set precede each
element in the validation set during each iteration.
This type of cross validation is important since we are using lagged features to predict
future transfer rates.
We measure the performance based on Root Mean Square Error (RMSE) in this process.

\begin{algorithm}
\begin{algorithmic}
\STATE Inputs: Training set ($X$), modeling method $M$, number of hyperparameters to try ($num\_params$), number of CV iterations ($k$), CV training and test widths ($train\_width$, $test\_width$), CV training and test set sizes ($train\_size$, $test\_size$)
\STATE Sort $X$ in increasing order of start time
\FOR{$i$ = 1 to $num\_params$}
\STATE Set $\alpha \leftarrow$ randomly sampled hyperparameters
\FOR{$j$ = 1 to $k$}
\STATE Set $train\_region \leftarrow$ random consecutive $train\_width$ rows of $X$
\STATE Set $train\_set \leftarrow$ random subset of $train\_region$ of size $train\_size$
\STATE Set $test\_region \leftarrow$ $test\_width$ rows of $X$ following $train\_region$
\STATE Set $test\_set \leftarrow$ random subset of $test\_region$ of size $test\_size$
\STATE Train model $M$ on $train\_region$ and evaluate performance (RMSE) on $test\_region$, where the prediction is $\max$(0, current prediction).
\ENDFOR
\IF{average of test RMSE's for $\alpha$ is lowest so far}
\STATE Set $\alpha_{best} \leftarrow \alpha$
\ENDIF
\ENDFOR
\RETURN $\alpha_{best}$
\caption{Nested Cross Validation that preserves the time order of events.}
\label{alg:cv}
\end{algorithmic}
\end{algorithm}

\paragraph{Baseline model with static features}
Now that we have selected the class of methods for constructing the
prediction models, the key next step is to choose the features to be
used for the models.
The first set of features contains 22 fields from the given performance
monitoring data set.
These features are known before the start of the file transfer and
called static features for the purpose of this modeling exercise.
These features include:
\begin{enumerate}
\item[A:] file size, instrument, experiment number, target
  host, target file system, and source storage system.
\end{enumerate}
Among these features, instrument, source file system and target file
system are encoded using one-hot encoding, and experiment number is
treated as categorical values.

With a good amount of exploration of the hyperparameter space and
different decision tree based modeling methods, the best of our
tree-based modeling approach turned out to be XGboot, and it gives a
RMSE of 64.3 MB/s when we use the first 90\% of the time series as
training and last 10\% of the time series for testing.

From the feature importance matrix in
Table~\ref{table:feature_importance_static}, which only displays the
5 most important features, we notice that file size is the dominant
factor affecting the transfer rate, and experiment number also has large
importance on the transfer rate.
The instruments, all seven of them together, count for approximately
10\%, which make it the third most
important in predicting transfer rates.  Among the instruments, cxi is the
most influential, presumably, it is the most frequently used instrument,
as shown in Table~\ref{table:instr_count}.

\begin{table}[tb!]
\centering
\begin{tabular}{|l|r|}
\hline
\textbf{Feature}                                             & \textbf{Importance} \\ \hline
File Size         & 66.827\%      \\ \hline
experiment number & 17.325\%      \\ \hline
cxi(instrument)   & 6.153\%      \\ \hline
ffb11(source file system)      & 4.855\%      \\ \hline
mfx(instrument)   & 2.604\%      \\ \hline
\end{tabular}
\vspace{0.2cm}
\caption{Feature Importance according to XGBoost
  with static features.}
\label{table:feature_importance_static}
\vspace{-4mm}
\end{table}

\section{Dynamic Features: Time Lags}
\label{sec:lags-trees}

The previous model with the static features uses only the information
from the current event at time $i$.
In statistical techniques for time series modeling, we often uses
information from time $i-1$, $i-2$, and so on~\cite{deGooijer:1992:TSM,
  Kane:2014:CAR}.
These most recent past data records are also called lags because they
have a fixed amount of lags in time (index) dimension.
We typically refer to the data record at time $i-1$ as lag 1 of the data
record at time $i$, similarly, the record at time $i-2$ is lag 2, and
record at time $i-\ell$ is lag $\ell$.
Next, we consider taking information from some of these lags to augment
the feature set used in our tree based models.
One important caveat to note here is that the most important piece of
information from these past events are their data transfer rates.  In
order for the data transfer rates to be available, these past events have
completed before the current event ($i$) starts.

The standard method of select information from the past is based on time
alone.  However, since our time series has a number of different
attributes/features, we can define different types of lags based on
the features other than time.
For example, since each LCLS experiment has its own specially image
capturing device and its own dedicated storage system, how quickly the
last completed file transfer would be strongly correlated with the speed
of current file transfer.  This is because the two file transfers share
many common systems.  The information from the recent past event, i.e.,
the lag, could be highly useful for predicting the performance of the
current file transfer.

After some exploration, we define five different types of lag variables.
Four of them share the same instrument, experiment, source file
system or target file system with the current event.
In other words, these lag variable are
generated by taking the value from the most recently finished job on the
same instrument, experiment, source file system or target file system.
Because instrument, experiment, source file system and target file
system are all known information from the data record at time $i$, the
most recently completed events with these information are easily
identifiable with suitable preprocessing.
The fifth one takes the value from the most recently finished job, i.e.,
considering the time alone.
There are some correlations between these lag variables.  For example, transfer rate from the most recently
finished job on the same instrument is highly likely to be the same as
that in the same experiment.

We are not only interested in lag 1, but also lags beyond the first one.


After an extensive amount of cross validation, we choose add to the
static features the following
set of dynamic features based on the lags with different properties.
\begin{enumerate}
\item[A:] file size, instrument, experiment number, target
  host, target file system, and source storage system.
\item[D1:] transfer rates and time differences of lag 1 on the same instrument,
  experiment, source file system or destination file system; transfer
  rate of lag 1 and lag 5 in time alone.
\end{enumerate}



\begin{figure}
    \centering
    \includegraphics[scale=0.27]{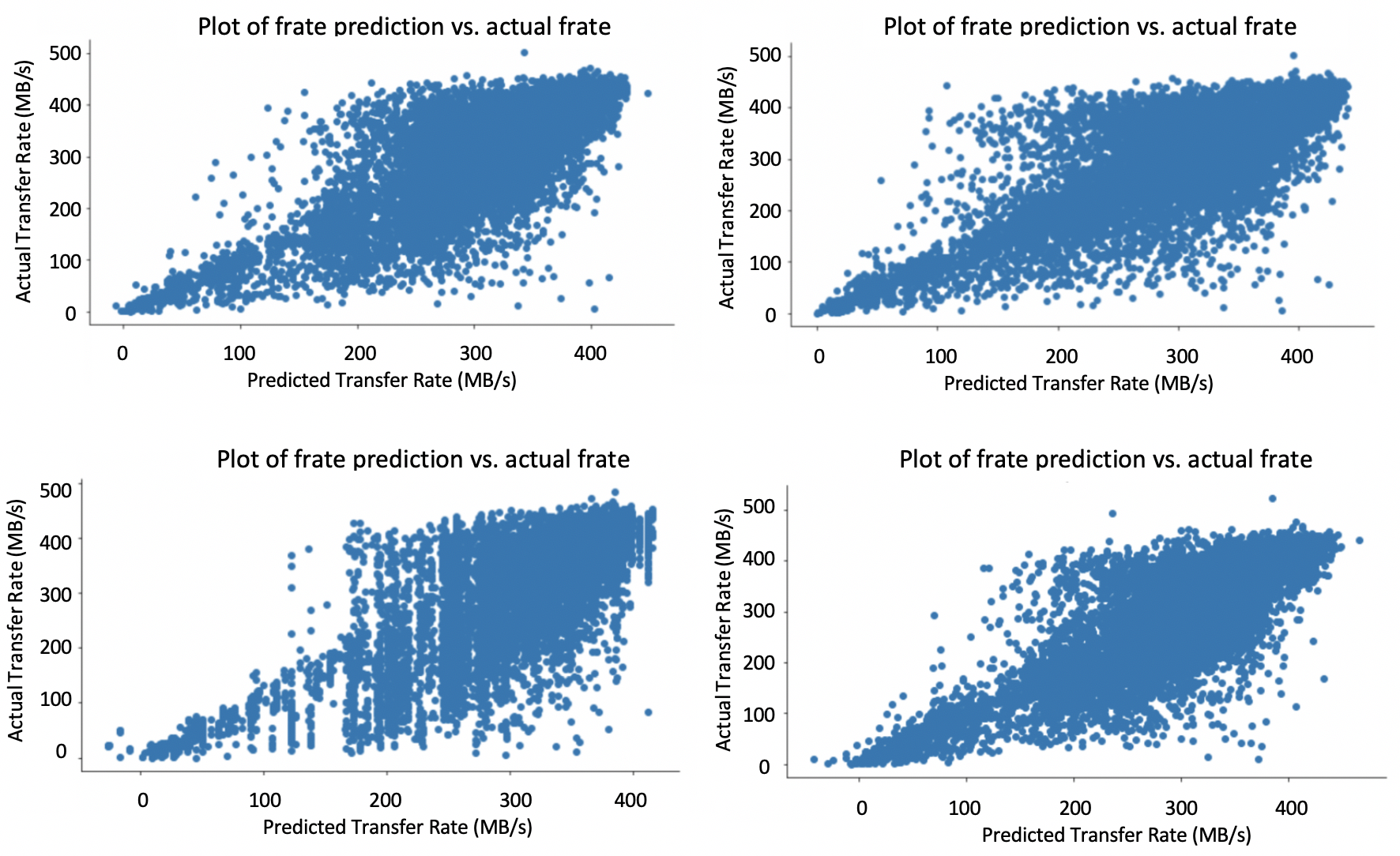}
    \caption{Predicted transfer rate vs.~Actual transfer rate using both
      Group A and Group D1 features (Top Left: Gradient Boosting (RMSE
      58.6), Top Right: Random Forest (optimized with nested cross
      validation, RMSE 60.0), Bottom Left: Random Forest (with default
      parameters), Bottom Right: Gradient Boosting (optimized with nest
      cross validation, RMSE 56.9).)}
    \label{fig:lags_predict_result}
\end{figure}

We have tested the performance of many combination decision trees and
their hyperparameter values.  
Figure~\ref{fig:lags_predict_result} shows a sample of the prediction
results plotted against the actual observed performance.
From this set of tests, we see that the Gradient Boosting method
achieved the smallest RMSE overall.  By
using the parameters: learning rate = 0.1, n estimators = 600, max
features = 4.12, max depth = 11, min samples split = 700, min samples
leaf = 10, the Gradient Boosting method achieved RMSE of 56.9MB/s, which
is about 12\% less than the RMSE achieved with the static features in
Group A.

The new feature importance for the tuned Gradient Boosting Tree is shown
in Table~\ref{table:feature_importance_gbt_cv}.
Comparing Table~\ref{table:feature_importance_static} with
Table~\ref{table:feature_importance_gbt_cv}, we can see the importance of
file size decreases with the lag variables, which indicates the lag
variables are indeed contributing to transfer rate prediction.  

In Tables~\ref{table:delayed1} and \ref{table:delayed2}, we showed two
cases where large gaps between start of what otherwise might be
consecutive files could indicate significant performance variations.
The presence of the variable ``time difference between same experiment''
in Table~\ref{table:feature_importance_gbt_cv} is thus expected.
However, the differences between Tables~\ref{table:delayed1} and
\ref{table:delayed2} suggest that further investigation is needed.

\begin{table}
\centering
\scalebox{0.9}{%
\begin{tabular}{|l|r|}
\hline
\textbf{Feature}                                             & \textbf{Importance} \\ \hline
File Size         & 26.868\%      \\ \hline
lag1 from same experiment - transfer rate & 22.614\%      \\ \hline
lag1 on same instrument - transfer rate & 13.642\%      \\ \hline
lag1 overall - file size   & 9.254\%   \\ \hline
lag5 overall - transfer rate   & 8.766\%   \\ \hline
experiment number & 6.496\%      \\ \hline
ffb11(srcfs)      & 3.599\%      \\ \hline
time difference between same experiment & 2.940\%      \\ \hline
cxi(instrument)   & 1.937\%      \\ \hline
mec(instrument)   & 0.830\%      \\ \hline
\end{tabular}}
\vspace{0.2cm}
\caption{Feature Importance of GBM with Time Difference Feature and Cross Validation}
\label{table:feature_importance_gbt_cv}
\vspace{-4mm}
\end{table}

\section{Automatic Feature Discovery with Neural Networks}
\label{sec:lags-nn}

From the feature engineering work presented in the previous sections, we
can confirm that constructing a complete feature set is a challenging and
tedious work.
For example, the dynamic features is Group B1 are chosen after pretty
extensive set of cross validation runs, however, there is no guarantee
that our exploration were complete in any way.
Inspired by a number of reported effort on using neural networks to
automate the feature selection process~\cite{Anderson:2016:ISF,
  Chen:2019:IPM, Fan:2019:DLF, Kang:2020:PRR, LeGuennec:2016:DAT}, we next proceed to explore how neural
networks might be used to reduce the effort of manual exploration.
Some of the key characteristics of neural network include its ability to
learn the complicated non-linear interactions among the input features
and to produce new features to represent these interactions.
In this work, we plan to explore three different neural network architectures
that extract different types of information from time series.
The first one only has fully connected layers, which is considered a general-purpose basic architecture~\cite{Demuth:2014:NND},
the second one includes convolutional layers, which use convolution to extract trends in time series~\cite{Lawrence:1997:CNN, LeGuennec:2016:DAT},
and
the last one uses a long short term memory (LSTM) model, which is a pure auto-regression process designed to model complex time series based~\cite{Hochreiter:1997:LSTM,Gers:2002:LSTM,Sundermeyer:2012:LSTM,Rangapuram:2018:DSS}.
In addition, we design a different selection of features from the current and past
data transfer events for each of the three different neural networks.
To take advantage of the neural networks as a feature generation
mechanism, we also explore combining the intermediate outputs from these
neural networks to generate an ensemble network.
Tests show that this ensemble network is able to achieve the best
prediction accuracy on average.

Next, we first describe the construction of these four neural networks
and their input features, and then describe how they perform.

\begin{figure}
    \centering
    \includegraphics[width=\linewidth]{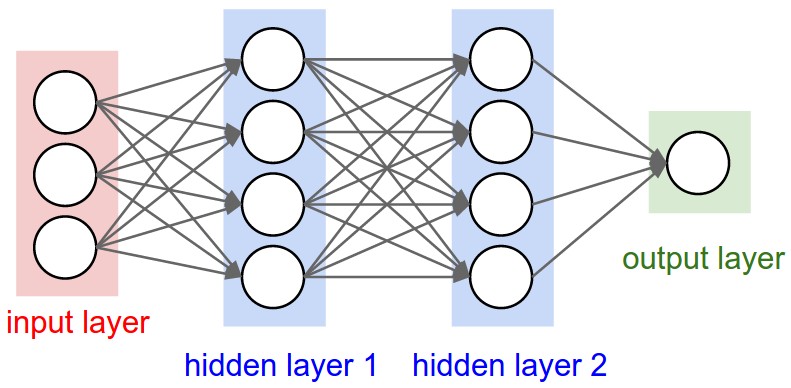}
    \caption{3 Fully Connected Layer Model}
    \label{fig:fc}
\end{figure}

\subsection{Neural Network Structures and Their Input Features}
The first neural network model implemented is a 3-layer fully connected
model, see Figure~\ref{fig:fc}.
In terms of input features for this neural network, we plan to use the
last set described in the proceeding section.  We add two
features about date and time as well as the concurrency levels of the
file transfer activities~\footnote{These features are not explicitly
  present in the original recorded values, but can be extracted through
  know procedures: for example the time of day and data of week can be
  extracted from the unix time stamp representing the start time, and
  the concurrencies could be computed by examining the start and end
  time values of file transfers.  Note that these procedures do not
  require any domain knowledge about the specific scientific workflow.}.
These features are selected due to their
potential to represent daily and weekly network traffic patterns as well
potential for network traffic congestions.
Altogether, we have the following set of features as input to the 
\begin{enumerate}
\item[A:] file size, instrument, experiment number, target
  host, target file system, and source storage system.
\item[B:] day of week, hour of day.
\item[C1:] number of active file transfers from the same
  experiment, number of active file transfers from the same instrument.
\item[D1:] transfer rates and time differences of Lag 1 on the same instrument,
  experiment, source file system or destination file system; transfer
  rate of lag 1 and Lag 5 in time alone.
\end{enumerate}

As in the previous models, the categorical variables are converted into
one-hot-encoding variables.

The second neural network architecture we explored is the Convolutional
Neural Network (CNN)~\cite{Lawrence:1997:CNN, LeGuennec:2016:DAT}.
CNN applies the mathematical operation known as convolution to extract features at different
scales.  This operation is very effective at discovering the local
information and relationship among the neighbor features.
In this particular application of predicting file transfer performance,
one critical information we need to infer is the status of the data
network carrying out the next file transfer.
As in most time series modeling tasks, we believe the state of the data
network could be approximated through the performance of the recent data
transfers.
We believe that it does not make sense to take a convolution over
different features, and plan to only apply convolution over time.
In order for the convolution operation to extract useful information we
need to provide enough information from the recent past.
We take 20 lag variables for each file transfer record, and each lag
variable is obtained from the most recently finished file transfer
belonging to the same experiment.
Altogether, the full list of input features includes:
\begin{enumerate}
\item[A:] file size, instrument, experiment number, target
  host, target file system, and source storage system.
\item[B:] day of week, hour of day.
\item[C1:] number of active file transfers from the same
  experiment, number of active file transfers from the same instrument.
\item[D2:] transfer rates of Lag 1--20 from the same
  experiment.
\end{enumerate}

\begin{figure}
    \centering
    \includegraphics[width=\linewidth]{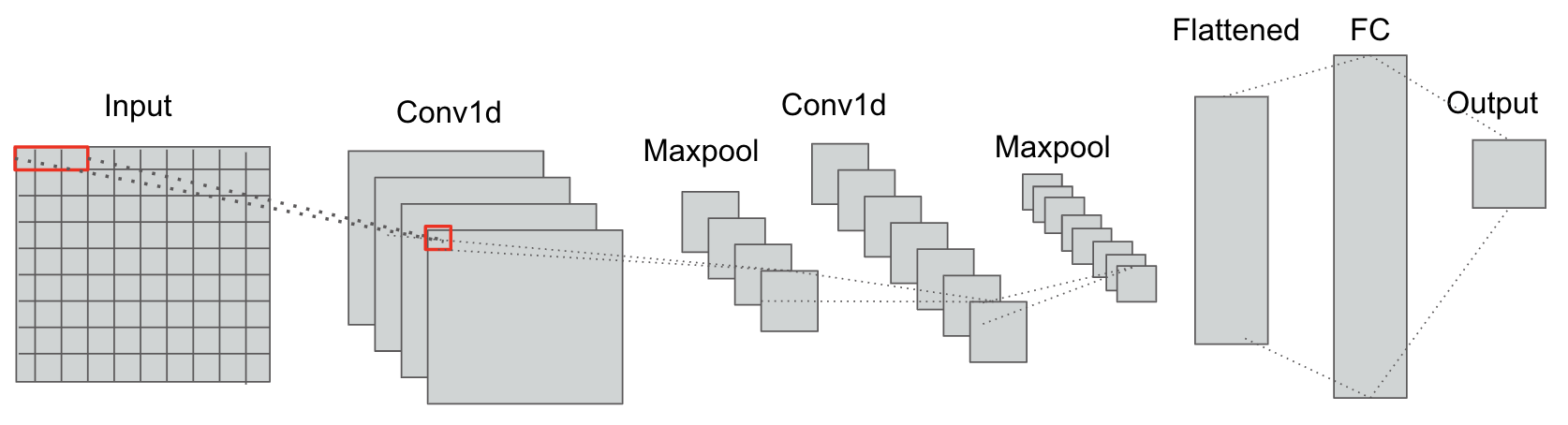}
    \caption{CNN Model}
    \label{fig:cnn}
\end{figure}

Following the convolution layer,
the activation function used is ReLU and a MaxPool layer is followed
after each CONV net.
The overall architecture of our CNN model is shown in Figure \ref{fig:cnn}.

The third neuroal network architecture we explore is Long Short-Term
Memory (LSTM)~\cite{Hochreiter:1997:LSTM}.
LSTM is a recurrent neural network that uses the information from the
current time step as well as recent past information to make
predictions.  This is especially well-suited for time series data
because it learns to store information over extended time intervals by
recurrent backpropagation.
We can regard LSTM as automatically discovering the dynamic lag
features, we only need to include the static features as input features.
The number of LSTM layers can be thought of the mechanism for
remembering past time steps.
In this case, we found two to four layers of LSTM work quite well.
In our tests, we found the five Lag 1 defined in Group D1 are useful as
input and we include all static features for each these five past events
along with their transfer rates and time differences mentioned in the
definition of Group D1.

If we only use the past information, then we will be missing the static
features of the current data transfer as input feature.
To make the data record about the current event match the structure of
the past events, we add the average of the transfer rates from the past
five events in Group D1 as the expected transfer rate for the current
event.
In later discussions, this set of features is referred to as the
\emph{projected performance}.

Following the LSTM layers, we have added one or more fully connected
layers before the final prediction.
In this regard, we can think of LSTM layers and the fully connected
layers as creating features for the final prediction.

\begin{figure}[H]
    \centering
    \includegraphics[width=\linewidth]{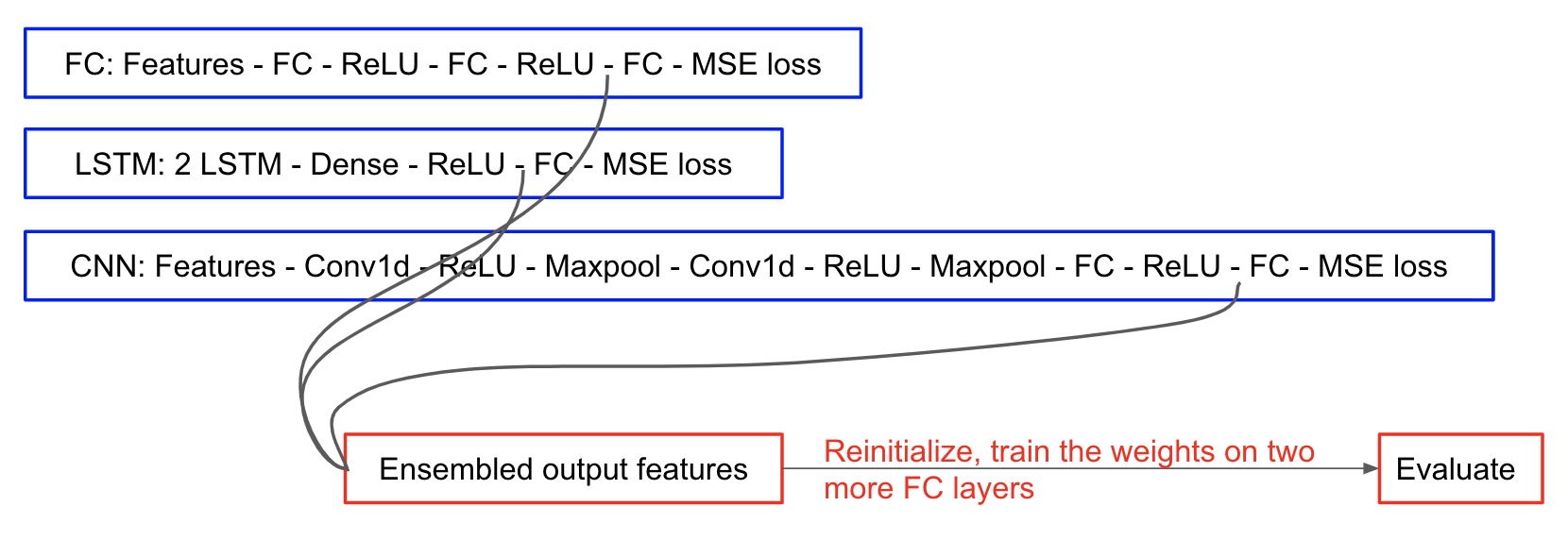}
    \caption{Ensemble three models}
    \label{fig:ensemble}
\end{figure}

The above three neural networks all ends with a fully connected layer to
produce the final output, we regard this final fully connected layer as
performing final combination of the intermediate features extracted by
the neural networks, and the input to this final layer as the final
features extracted.
The fourth model is an ensemble model that makes use of all these final
intermediate features.
We have conducted a number of different tests on combining the models, 
and eventually choose to use the combination shown in Figure \ref{fig:ensemble}.



To select the hyperparameters for the neural networks mentioned above,
we use Adam~\cite{Kingma:2014:Adam} with mean square error (MSE) of
predictions as the loss function.
Each of the four models are tuned separately.

\subsection{Test Results with Neural Network Models}
In our experiments with the fully connected model, we have explored a
number of different variations by adding regularization, dropout layer
and step-dependent learning rate.
Our tests show that the L2 regularization does not reduce the average
prediction error with the 3-layer fully connected model.
Similarly, the strategy of dropping out some links also increases
prediction error, most likely because our model is relatively simple and
there is no links to drop.
We find adjusting the learning rate based on the step count could reduce
the average prediction error noticeable.

After extensive cross validation and optimization with Adam, we find
that a learning rate of $10^{-3}$ and hidden layer size of 250 (for both
hidden layers) give the best validation loss.
We adjust the learning as follows: (1) start with the initial learning rate
($10^{-3}$); (2) if the loss has not decreased in 40 epochs, then reduce the
learning rate by a factor of $0.1$; (3) use the same learning rate for
no more than 120 epochs before reducing it by $0.1$.
Applying this model to the test dataset, we obatin a RMSE of 71.48MB/s.


With the CNN model, we similarly experimented with different learning
rate, hidden layer size, filter size, convolution window size and so on.
The smallest loss is achieve with learning rate = 1e-3, hidden layer
size = 128, filter$\_$size1 = 3, filter$\_$size2 = 3, filter$\_$num1 =
32, filter$\_$num2 = 16.
The resulting RMSE on test data is 76.01MB/s.

For the LSTM model, there are also many ways we can tweak the model and
architecture.  We have presented some of the best results in Table~\ref{table:lstm},
indicate that using a relatively larger hidden size of LSTM layer can increase the model complexity and encourage the model to learn more from the features.  Beside of that, adding the static features of the current transfer can increase the prediction accuracy.

\begin{table*}
\centering
 \begin{adjustbox}{width=\columnwidth*2,center}
\begin{tabular}{|l|l|r|r|}
\hline
\textbf{Input Feature} & \textbf{Layers} & \textbf{Hidden Size of LSTM layer} & \textbf{RMSE (MB/s)} \\ \hline
Group D1 + static features & 2 LSTM - 1 FC & 3 & 79 \\ \hline
Group D1 + static features & 4 LSTM - 1 FC & 3 & 85 \\ \hline
Group D1 + static features & 2 LSTM - 1 FC & 8 & 77   \\ \hline
Group D1 + static features + projected performance & 2 LSTM - 1 FC & 8 & 73     \\ \hline
Group D1 + static features + projected performance & 2 LSTM - 1 FC & 16 & 72     \\ \hline
Group D1 + static features + projected performance & 2 LSTM - FC - ReLU - FC & 16 & 66   \\ \hline
\end{tabular}
 \end{adjustbox}
\vspace{0.2cm}
\caption{Results from different LSTM networks}
\label{table:lstm}
\end{table*}

We finally take the best hyper-parameters and network architectures from
each of the model to form an ensemble model, see
Figure~\ref{fig:ensemble}.  We take the intermediate feature outputs
after the fully connected layer in each model, and ensemble the feature
outputs together to form the new feature set.  The ensemble features are
then trained through a two-layer fully connected model.  After some
additional optimization with Adam, this ensemble model achieves RMSE of 59.8MB/s.

\section{Dynamic Features: Incorporating Data Management Details}
\label{sec:runs}

Section~\ref{sec:testcase} has a description of the variables captured
by the performance monitoring system at LCLS.  In the previous modeling
exercises, these variables are used
as static features labeled as Group A.  There are two exceptions.
The first exception is the unix time stamp for start time\footnote{The
  end time is considered as unknown or to be predicted, therefore, not
  usable as input to the modeling procedures.}, which is digested into a
couple of different features in Group B used in the previous models.
The second exception is the file name of the data file being
transferred, which has not been used so far.
Since the file name is unique for each file, we were not sure how to
make use of it until we notice that the file name is composed of the
experiment, chunk, stream, and a sequence number with the stream, see
more discussion in Section~\ref{sec:testcase}.
The data acquisition system of LCLS composes these file names use a
numerical representation of these properties about the
physical experiment.
Even though such information is specific to LCLS, however, we can
imagine that the automated data collection system would similarly
compose data file names based on unique attributes of the data
collection process so as to ensure the file names are unique.

Among the features that could be extracted from the file names, the
experiment number and the chunk number are already explicitly recorded
by the performance monitoring system.
The new features are the stream number and the sequence number (within a stream).
Next, we describe our attempt to make use these new features to further
improve the prediction accuracy.

Based on the previous experiences
and with some additional explorations, we decided to expand the list of
features as follows:
(1) Expand Group C1 with additional measures of concurrency for taget
storage systems to form a new Group C2.  These features should better
reflect the workload on the target storage systems.
(2) Expand Group D1 with additional information both on the source
storage system as well as the target of the file transfer to form a new
Group D3.  This change is designed to capture the status of both
storage systems involved in a file transfer.  Through extensive
testing, we decided to only include information from the most recently
completed file transfers with the same experiment number and so on.
This is an expansion of the previous defined Group D1, with fewer lags
than used in Group D2.\footnote{We were hoping LSTM would automatically
  identify the most important features from Group D2,
Unfortunately, additional exploration is needed to ensure this automatic
process could produce more accurate models.}
(3) Based on the examples shown in Tables~\ref{table:delayed1} and
\ref{table:delayed2}, we expect the time differences between start of a
file transfer in a stream has significant impact on the actual data
transfer rate, even if the mechanisms for the relation is not
obviously visible.  Such information could be extracted once we figure
out how to digest the file names to identify the streams and sequence
within a stream.  This time difference within a stream is captured in a
new feature group E.

In summary, we have the the following feature
set as input to next prediction models:
\begin{enumerate}
\item[A:] file size, instrument, experiment number, target
  host, target file system, and source storage system.
\item[B:] day of week, hour of day.
\item[C2:] Number of total jobs and number of unique experiments running on
  the same target file system, target host, and node, respectively, when the current job
  is started.
\item[D3:] Statistics (transfer rate, file size, time between last job's stop
   time and current job's start time) from the last job on the same
   instrument, target file system, target host, target node, experiment,
   and chunk, respectively.
 \item[E:] Time difference between the start time of the first job of the
   chunk and the start time of the current job.
\end{enumerate}

\paragraph{Random Forest Results}
Using the nested CV algorithm~\ref{alg:cv} with the training set $X$
equal to the first 90\% of the rows, $k=10$, $train\_width = 20000$,
$train\_size = 5000$, $test\_width = 2000$, $test\_size = 500$, we then
retrain the random forest model using the best hyperparameters on a
30000 row subset of the first 90\% of the rows of $X$, and evaluate RMSE
on a 3000 row subset of the last 10\% of the rows of $X$, where the
prediction is $\max$(0, prediction).  This yields an RMSE of 52.8 MB/s.

\begin{figure}
    \centering
    \includegraphics[width=3.25in]{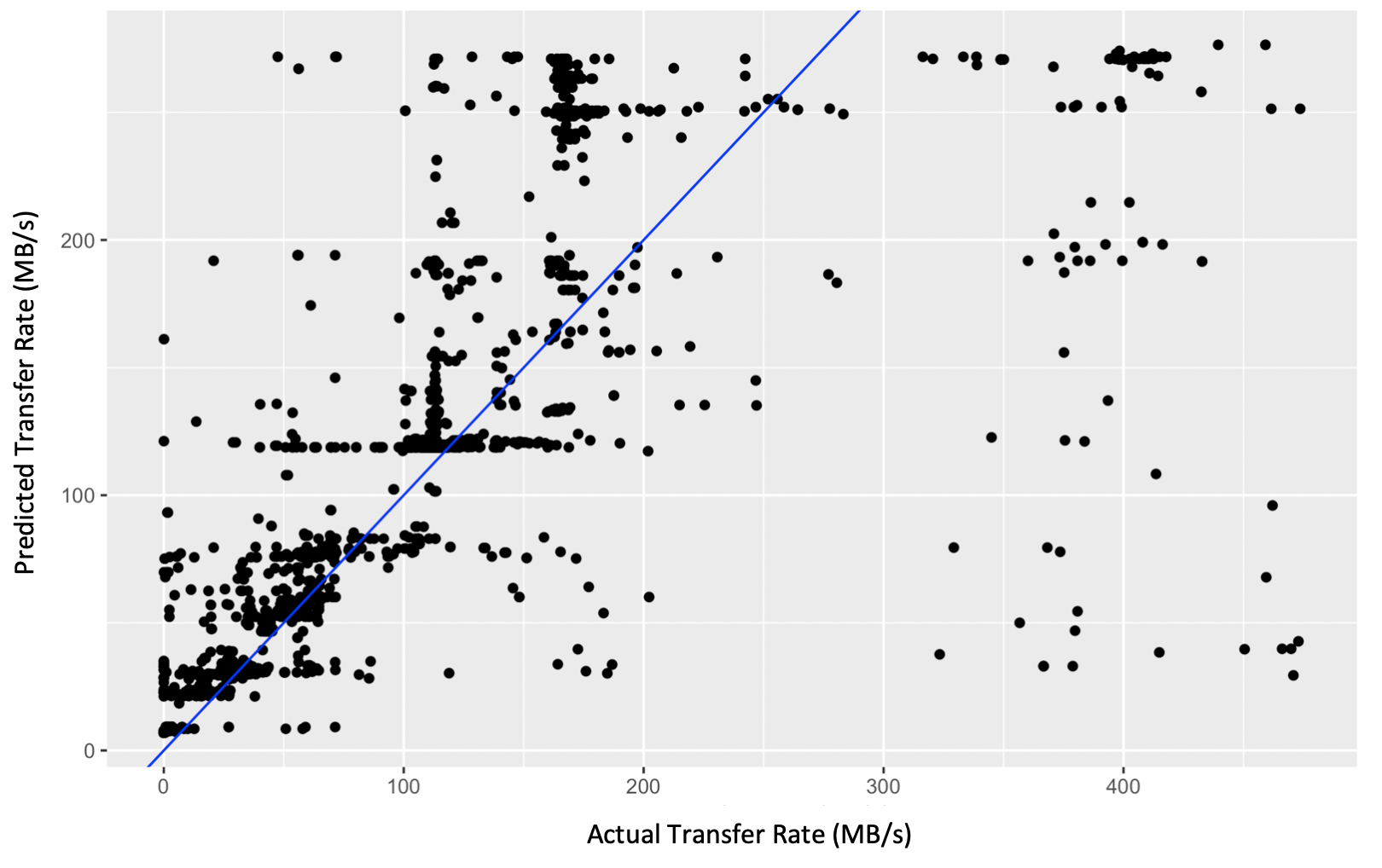}
    \caption{Actual vs Predicted plot using Random Forest}
    \label{fig:df-actual-pred}
\end{figure}

A scatter plot of the actual versus predicted transfer rates is shown in Figure \ref{fig:df-actual-pred},
and the most important features are given in Table~\ref{tb:RF-feat}.

\begin{table}
\centering
\begin{tabular}{|l|r|}
\hline
Feature					& \textbf{Importance} \\ \hline
Lag 1 same Instrument, Transfer Rate	& 70.4\%  \\ \hline
Lag 1 same File System, Transfer Rate	& 7.9\%   \\ \hline
Lag 1 same Node, Transfer Rate 		& 7.0\%   \\ \hline
File Size				& 4.1\%   \\ \hline
Lag 1 same Instrument, Time Diff	& 4.1\%   \\ \hline
\end{tabular}
\vspace{0.2cm}
\caption{The top five most important features for the Random Forest
  model.}
\label{tb:RF-feat}
\end{table}

\paragraph{Xgboost Results}
Using the nested CV algorithm~\ref{alg:cv} with the training set $X$
equal to the first 90\% of the rows, $k=10$, $train\_width = 20000$,
$train\_size = 6000$, $test\_width = 2000$, $test\_size = 800$, we then
retrain the Xgboost model using the best hyperparameters on the same
30,000 row subset as the random forest, and evaluate the RMSE on the
same 3000 row subset as the random forest, where again the prediction is
$\max$(0, Xgboost prediction).  This yields an RMSE of 36.1 MB/s.

A scatter plot of the actual versus predicted transfer rates is shown in Figure~\ref{fig:actual-pred},
and the most important features for this model is listed in Table~\ref{tb:XGB-feat}:
\begin{figure}
    \centering
    \includegraphics[width=3.25in]{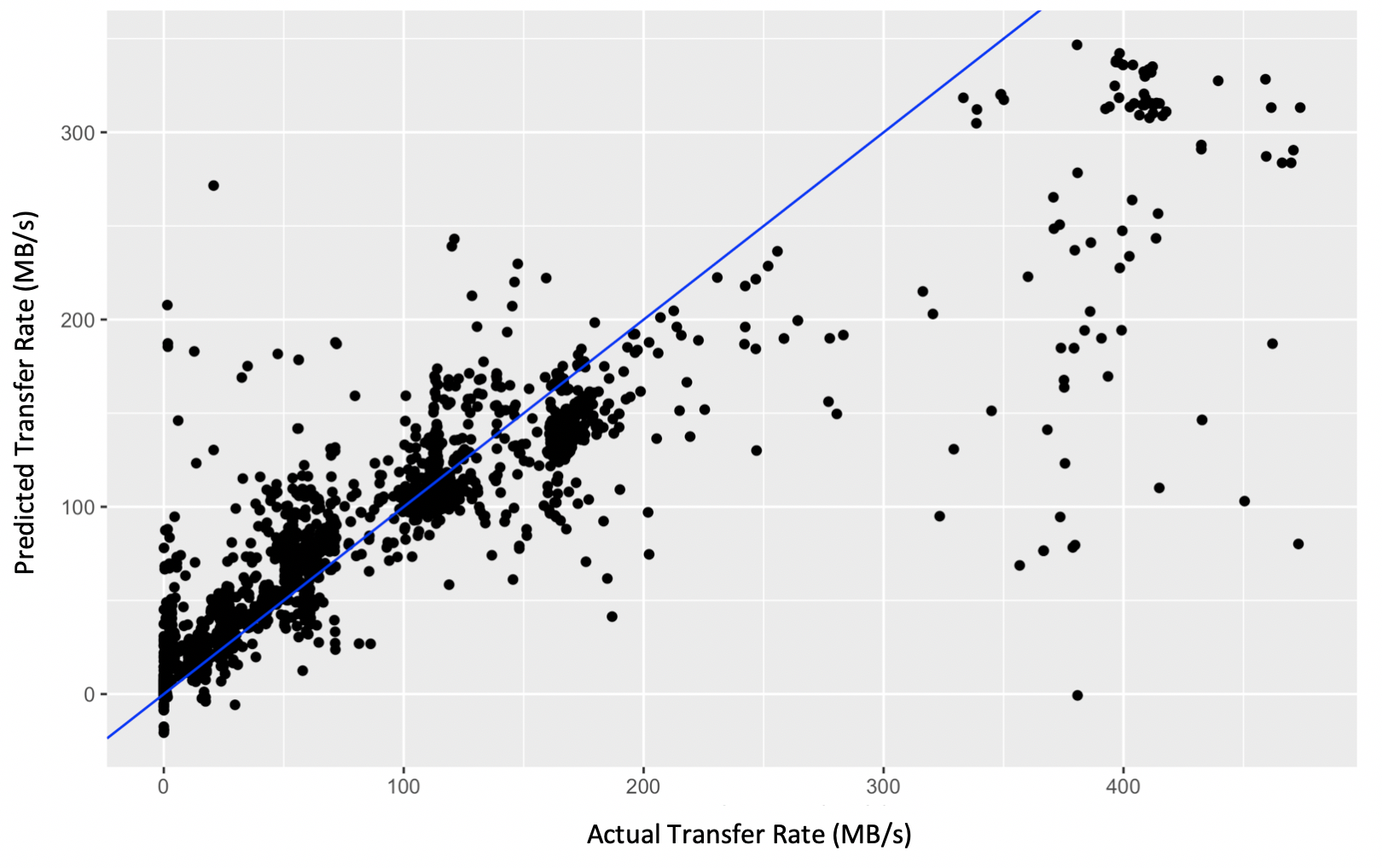}
    \caption{Actual vs Predicted plot using Xgboost}
    \label{fig:actual-pred}
\end{figure}    

\begin{table}
\centering
\scalebox{0.85}{%
\begin{tabular}{|l|r|} \hline
Feature                                         & \textbf{Importance} \\ \hline
Instrument					& 12.9\%     \\ \hline
Lag 1 same Experiment, File Size		& 9.5\%      \\ \hline
Day of the Week					& 7.9\%      \\ \hline
Lag 1 same Experiment, Time Diff.		& 7.5\%      \\ \hline
Target Host					& 6.2\%      \\ \hline
Lag 1 same Target File System, Transfer Rate	& 5.3\%      \\ \hline
Lag 1 same Node, Time Diff.			& 4.4\%      \\ \hline
Lag 1 same Instrument, Transfer Rate		& 3.9\%      \\ \hline
\end{tabular}}
\vspace{0.2cm}
\caption{The features with more than 1\% importance according the
  XGBoost procedure used for Figure~\ref{fig:actual-pred}.}
\label{tb:XGB-feat}
\end{table}

Recall that the most accurate model with the static features in the original
performance monitoring data (i.e., Group A) achieved an RMSE of 64.3
MB/s.  The newly expanded feature set developed in this section produces
clearly more accurate prediction models, where the best RMSE achieved is
36.1 MB/s.
From the scatter plots in Figures~\ref{fig:lags_predict_result}, \ref{fig:df-actual-pred}, and \ref{fig:actual-pred},
we could also easily see that the expanded feature set reduces the
differences between predictions and actual measurements.
The points in Xgboost are more aligned with the y = x line.

\paragraph{Observations}
On the question of which features are the most important for the
prediction models, we see one common trend from
Tables~\ref{table:feature_importance_static},
\ref{table:feature_importance_gbt_cv}, \ref{tb:RF-feat}, and
\ref{tb:XGB-feat}, that is, the performance of the most recently
completed data transfers is highly influential in predicting the current
file transfer performance, especially the last event, also known as Lag 1, on the same data
path, because these data transfer events share the same source storage
system, the same communication network, and the same target storage
system.
When the Lag 1 has just completed recently, we can expect the current
file transfer to have the same performance, because the current state of
the systems involved in the data transfer must be similar to that the
just completed one.
Given this general observation, the key objective of our feature
engineering task would be to discover the best way to represent the
current status of the system by teasing out condition that is most
similar to the current event.  This is a general lesson that is
applicable to many application scenarios.

Even though we started out this section with the extraction of
application specific information from file names, i.e., the stream
number and the sequence number within a stream, however, the feature
importance results from the actual modeling exercises, see
Tables~\ref{tb:RF-feat} and \ref{tb:XGB-feat}, indicate that these
application specific features are not particularly important, none of
the application specific features made to the top of the feature
importance lists.  Therefore, we may not need to find the most
perfect match when identifying the lag variable.
We would have expected the more accurately a lag event matches with the
current event the more similar the performance, however, the actually
underlying systems are influenced by more variables than is captured by
the monitoring system, therefore, even if we can match all the features
we know of, there is no guarantee that the two events would actually be
perfectly matched.
Furthermore, the more specific is the matching condition, the fewer
events would satisfy, which would lead to more records with no matching
past event or stale matches from far past where the states of the
systems are not comparable any more.

We also see significant difference among the feature importance tables,
which suggests that we have yet found a definitive set of features that
would best model the particular time series.
Additional work is need to verify whether there are better ways to
construct a feature set.
For the moment, we advise readers to include more variety of features.
Since the models including feature group D3 achieved lower RMSE,
we believe the feature group D3 is better than D1 and D2.
Even the group D2 contains more variables, the group D3 has more
variety, which we believe is key to its success.

\section{Summary}
\label{sec:summary}
When analyzing time series data, we often face the situation where some
features related to the key objective are missing from
the existing measurements.
For example, in our task of predicting the file transfer performance, we
would ideal know exactly how busy are the source storage system, the
data communication network, and the target storage system, however, such
information is missing from the performance monitering data we have
access to.
In this situation, we could extract
proxy information from the recent past records.
In this work, we
explore three approaches to extract such dynamic information from a set
of network data transfers from a large experimental facility and data
center.
We first try to use a fixed number of recent past records
(also called lags), an approach is applicable to any time series data.
We find that these dynamic features are able to reduce the average prediction
error by about 12\%, using the same Gradient Boosting Trees that was
also used to model the data transfer performance with only static
features in the raw data.
Our second approach utilizes neural networks of various configurations to
automatically extract key features from recent past data records.  Even
though this approach was successfully used in the literature, we were unable to find a
neural network that was more accurate than the Gradient Boosting Trees.
Finally, we explored an application specific technique that took
advantage of the information embedded in the file names to identify
recent past data records involving the same compute, storage, and
network systems.
This set of application specific features allowed us to build a more
accurate Gradient Boosting Trees, where the prediction error was reduced
to 36.1, a 44\% reduction compared with the original model with static
features.

Since our approach augment the time series data with information from
the past, it is critical that we avoid leaking information among the data sets
for training, testing, and validation.  In this work, we
designed a nest cross validation approach that guarantees to prevent
this leakage.  We believe that this algorithm is key to selecting the
approach hyperparameters for our machine learning models.

For future studies, we plan to study the propose nest cross validation
algorithm more carefully.
Even though we believed the application specific information was
critical to the success of the final feature set presented in
Section~\ref{sec:runs}, the feature importance information from the
actual models does not agree with our expectation.
It would be worthwhile to examine this discrepancy further.
In addition, we believe the neural network approach has great potential
despite our current inability to make an effective use.
For example, our current training data range might be too narrow or too
wide, by adjusting this training data range, we might be able to capture
the most relevant trends and improve the prediction accuracy.

\section*{Acknowledgment}
This work was supported by the Office of Advanced Scientific Computing Research, Office of Science, of the U.S. Department of Energy under Contract No. DE-AC02-05CH11231, and also used resources of the National Energy Research Scientific Computing Center (NERSC). 
Use of the Linac Coherent Light Source (LCLS), SLAC National Accelerator Laboratory, is supported by the U.S. Department of Energy, Office of Science, Office of Basic Energy Sciences under Contract No. DE-AC02-76SF00515. 
The authors would like to thank Kari Hanson and Professor Mary Wootters of Stanford University for setting up and sponsoring this project as part of ICME's Xplore project.

\bibliographystyle{plain} 
\bibliography{biblio}


\end{document}